\documentclass[runningheads]{llncs}
\usepackage{graphicx}
\usepackage{amsmath,amssymb} 
\usepackage{graphics,subfigure}
\usepackage{color}
\usepackage{makecell}
\usepackage[width=122mm,left=12mm,paperwidth=146mm,height=193mm,top=12mm,paperheight=217mm]{geometry}

\begin{document}


\mainmatter

\title{Uncorrelated Feature Encoding \\ for Faster Image Style Transfer} 

\titlerunning{ }

\authorrunning{ }

\author{Minseong Kim$^{1}$, Jongju Shin$^{2}$, Myung-Cheol Roh$^{2}$, Hyun-Chul Choi$^{1}$}
\institute{	$^{1}$Yeungnam University, $^{2}$Kakaocorp \\ 
	\texttt{ \{tyui592, pogary\}@ynu.ac.kr, \{isaac.shin, joshua.roh\}@kakaocorp.com } }

\maketitle

\begin{abstract}
Recent fast style transfer methods use a pre-trained convolutional neural network as a feature encoder and a perceptual loss network. Although the pre-trained network is used to generate responses of receptive fields effective for representing style and content of image, it is not optimized for image style transfer but rather for image classification. Furthermore, it also requires a time-consuming and correlation-considering feature alignment process for image style transfer because of its inter-channel correlation. In this paper, we propose an end-to-end learning method which optimizes an encoder/decoder network for the purpose of style transfer as well as relieves the feature alignment complexity from considering inter-channel correlation. We used uncorrelation loss, i.e., the total correlation coefficient between the responses of different encoder channels, with style and content losses for training style transfer network. This makes the encoder network to be trained to generate inter-channel uncorrelated features and to be optimized for the task of image style transfer which maintained the quality of image style only with a light-weighted and correlation-unaware feature alignment process. Moreover, our method drastically reduced redundant channels of the encoded feature and this resulted in the efficient size of structure of network and faster forward processing speed. Our method can also be applied to cascade network scheme for multiple scaled style transferring and allows user-control of style strength by using a content-style trade-off parameter.

\keywords{Image style transfer, Encoder learning, Uncorrelated feature encoding, Uncorrelation loss}
\end{abstract}

\section{Introduction}

\begin{figure}[t]
\begin{center}
\includegraphics[trim={2.8cm 5cm 5cm 3cm},clip,width=1.0\linewidth]{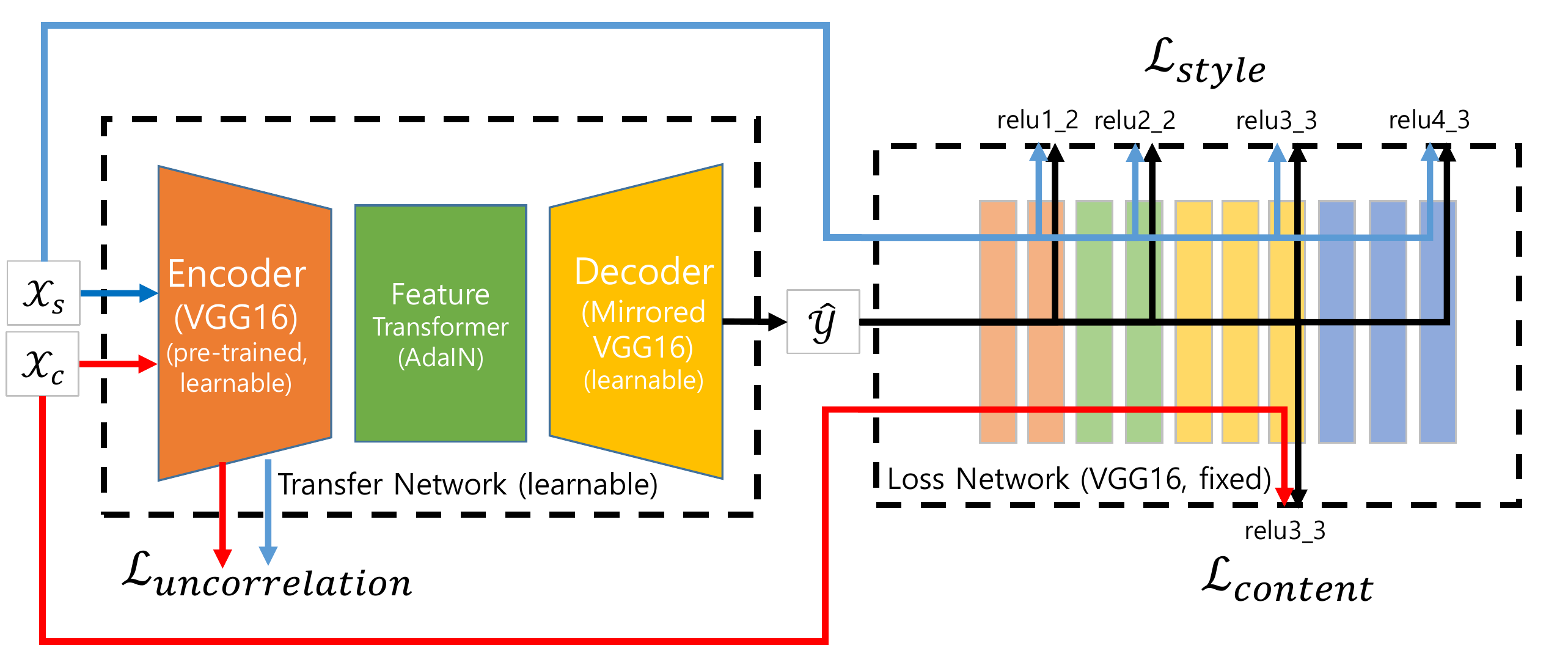}
\end{center}
   \caption{The overall configuration of our method. Similar to \cite{Huang_2017_ICCV}, the pre-trained VGG16\cite{Simonyan14c} feature network is used as the initial encoder, and the decoder network is configured to be mirrored structure of the encoder network. The correlation loss is calculated using the feature maps extracted from the content and style images, and the content and style losses are calculated using the pre-trained VGG16 feature network as \cite{Johnson_2016_ECCV}. As shown in the above diagram, decoder networks are learned during the training process but the loss network is fixed. We uses the feature alignment method using mean and standard deviation as \cite{Huang_2017_ICCV} in the feature transformer.}
\label{fig:network_architecture}
\end{figure}

Image style transfer is a computer vision application which synthesizes an output stylized image similar to a target image in style and to a source content image in semantic content. There are three major tasks in image style transfer, i.e., how to define content and style of an image, how to generate an image from the defined style and content, and how to transfer the style of the target image to the output image maintaining its content same to the source content image. For the first task, in the first neural approach\cite{Gatys_2016_CVPR}, the style of an image is defined as Gram matrices, i.e., the correlations of feature maps extracted from several layers of VGG\cite{Simonyan14c} feature network, and the content as a feature map from a higher layer of VGG feature network. These definitions of style and content are continuosly used with slight modification in the subsequent researchs of image style transfer. For the second task, online iterative pixel-wise optimization\cite{Gatys_2016_CVPR} and offline feed-forward network learning\cite{Johnson_2016_ECCV,Ulyanov_2016_ICML,Dumoulin_2017_ICLR,Huang_2017_ICCV,Li_2017_NIPS} methods have been proposed. The feed-forward network approaches speeded up image generating process to subrealtime by adopting encoder/decoder network architecture and changing the online optimization process of \cite{Gatys_2016_CVPR} to offine network learning with the objective function similar to \cite{Gatys_2016_CVPR}. For the last task of style transfer, modified instance normalization methods\cite{Dumoulin_2017_ICLR,Ghiasi_2017_BMVC,Huang_2017_ICCV} have been proposed to embed multiple or arbitrary styles in a feed-forward network. They interpreted the process of style transfer as an encoded feature distribution alignment under the assumptions of channel-uncorrelated Gaussian distribution\cite{Dumoulin_2017_ICLR,Ghiasi_2017_BMVC,Huang_2017_ICCV}. Later, a new alignment technique of channel-correlated Gaussian distribution\cite{Li_2017_NIPS} improved the generated style quality by using feature covariance with cascade network scheme.

However, the encoder/decoder network approach with correlation-considering alignment of feature distribution\cite{Li_2017_NIPS} still has some limitation. First, computing the square root or inverse of covariance matrix in forward process requires high computational cost. Second, VGG feature network in the previous feed-forward network methods\cite{Huang_2017_ICCV,Li_2017_NIPS} is not an optimized encoder for style transferring task but for image classification task. Third, the architecture of encoder/decoder network is also not optimized in structure.

In this paper, we introduce a simple but powerful method for end-to-end learning of feed-forward network to improve image style transfer in those three aspect of view. We define uncorrelation loss of encoded features and use it to train the encoder network to generate channel-uncorrelated features. Once the inter-channel correlation of the encoded features is removed, we need not use the channel-correlated feature alignment\cite{Li_2017_NIPS} anymore and can use the channel-wise independent feature aligment\cite{Huang_2017_ICCV} of much less computational cost. And not only decoder network but also encoder network is optimized to the image style transfer task by doing end-to-end learning with style and content losses to train encoder/decoder networks simultaneously. Moreover, the encoder network trained to minimize our uncorrelation loss is experimentally proven to achieve drastical reduction in dimension of the encoded features. This results in a compact network by removing redundant channels in convolution layer of decoder network. As the previous approaches, our method also allows user control of style strength\cite{Dumoulin_2017_ICLR,Ghiasi_2017_BMVC,Huang_2017_ICCV,Li_2017_NIPS} at runtime and can be applied to the cascade network scheme\cite{Li_2017_NIPS} improve style quality to multiple scaled styles.

The remained of this paper consist of four parts. In the part of related work, we will review previous researches which are related to image style transfer and motivated our methods. In the part of method, our uncorrelation loss and end-to-end learning method with or without cascade network scheme will be described. We will verify the effectiveness of our method based on a generalized experimental results in the part of experimental result, and will conclude this work in the last part.

\section{Related Work}

Gatys et al.\cite{Gatys_2016_CVPR} proposed a seminal work on image style transfer by adopting VGG\cite{Simonyan14c} feature network, a convolutional neural network trained for image classification. They used the correlation matrix, a.k.a. Gram Matrix, of the feature map extracted from low to high layers of VGG feature network as the style feature of image and the feature map from a higher layer of VGG feature network as the content feature of image. And they iteratively updated each pixel of an initial noisy image to minimize style loss, i.e., an error between the styles of the output image and target style image, and content loss, i.e., an error between the contents of the output image and the source input image, simultaneously by using a gradient-based optimization method. Although they achieved style transferring to arbitrary target style, their method has a pracitcal problem of very slow image generating speed because of their pixel-wise online optimization process.

Style transferring based on a feed-forward network\cite{Johnson_2016_ECCV,Ulyanov_2016_ICML} solved the problem of slow image generation by moving online iterative optimization to offline learning of feed-forward network with the same content and style losses in \cite{Gatys_2016_CVPR}. Additionaly, Ulyanov et al.\cite{Ulyanov_2017_CVPR} replaced batch normalization with instance normalization and this resulted in output style quality improvement. However, those methods based on a feed-forward network\cite{Johnson_2016_ECCV,Ulyanov_2016_ICML,Ulyanov_2017_CVPR} have a limitation of just one style per network.

Several methods have been proposed to deal with diverse tasks such as spatial-color control \cite{Gatys_2017_CVPR}, multiple style interpolation \cite{Dumoulin_2017_ICLR,Ghiasi_2017_BMVC,Huang_2017_ICCV} and photorealism \cite{Luan_2017_CVPR} in style transfer.

Dumoulin et al.\cite{Dumoulin_2017_ICLR} proposed conditional instance normalization (CIN) method which can synthesizes multiple styles in a feed-forward network by learning different affine parameters for each style in training phase and using the affine parameters corresponding to the target styles to transform input content feature to the style feature domain in test phase. Although they achieved multiple styles in a network, their method has a limitation in number of styles in a network and requires an additional training process for unseen styles. Another domain adaptation approach\cite{Li_2017_IJCAI} interpreted style transferring task as a domain adaptation problem with maximum mean discrepancy(MMD) with the second order polynomial kernel. Based on this interpretation, Huang et al.\cite{Huang_2017_ICCV} proposed adaptive instance normalization (AdaIN) method which transforms content image to arbitrary target style with a feed-forward network. They adjusted the feature distribution of content image into that of target style image by using means and standard deviations directly calculated from the features of content and style images. Their method assumed that the feature distributions are uncorrelated between channels in their AdaIN layer and style loss calculation, and this caused the lower quality of output style than that of previous approaches\cite{Gatys_2016_CVPR,Johnson_2016_ECCV,Ulyanov_2016_ICML,Ulyanov_2017_CVPR}. 

Sun et al.\cite{Sun_2016_AAAI,Sun_2016_ECCV} showed that aligning second order statistics of deep feature is useful to adapt a trained network from source domain to target domain. In the same manner, the latest domain adaptation approach\cite{Li_2017_NIPS} for image style transfer proposed whitening and coloring (WCT) method for distribution adjustment using means and covariance matrices of the features of content and style images. WCT improved the quality of transferred style by considering inter-channel correlation of features with more computational cost of calculating inverse square root of covariance matrices in forward process of network.

\section{Method}

Here, we describe how to remove the inter-channel correlation from the encoded feature while maintaining its semantic representational ability for image style transfer during end-to-end learning of a feed-forward network.

\subsection{Uncorrelation Loss}

Pearson's correlation coefficient\cite{Sedgwicke4483} is a popular measure of correlation between two random variables. If $c1$ and $c2$ are random variables, then their correlation coefficient is represented as Eq.\ref{eq:pearson's correlation coefficient}.
\begin{equation} \label{eq:pearson's correlation coefficient}
r = \frac{cov(c1, c2)}{\sqrt{var(c1)} \sqrt{var(c2)}}
\end{equation}
In the same manner, if we have an encoded feature $F = [f_1, f_2, ..., f_C] \in R^{C\times (H \times W)}$ of $C$ channel vectors of ($H \times W$) length, we can calculate inter-channel correlation coefficient $r_c$ between two channel vectors of the feature $f_{i}, f_{j} \in R^{(H \times W)}$ as Eq.\ref{eq:cosine_distance} because each channel data of the feature can be interpreted as a sample sets of each channel random variable.

\begin{equation} \label{eq:cosine_distance}
r_c(f_{i}, f_{j}) = \frac{(f_{i} - E(f_{i}))^T \cdot (f_{j} - E(f_{j}))}{\left \| f_{i} - E(f_{i}) \right \|\left \| f_{j} - E(f_{j}) \right \|}
\end{equation}
Then, we can define uncorrelation loss (Eq.\ref{eq:uncorrelation_loss}) as the summation of all the correlation coefficients between different channel vectors of the encoded features, $F_c = [{f_c}_{1}, {f_c}_2, ..., {f_c}_C]$ of content image and $F_s = [{f_s}_1, {f_s}_2, ..., {f_s}_C]$ of target style image.
\begin{equation} \label{eq:uncorrelation_loss}
\mathcal{L}_{uncorrelation} = \frac{1}{2} ( \sum_{i \neq j} r_c({f_c}_i, {f_c}_j) + \sum_{i \neq j} r_c({f_s}_i, {f_s}_j) )
\end{equation}

The inter-channel correlation coefficient of Eq.\ref{eq:cosine_distance} is geometrically equivalent to cosine distance between two channel residual vectors. Therefore, minimizing the uncorrelation loss of Eq.\ref{eq:uncorrelation_loss} is equivalent to learning the orthogonal independent basis of channels.

\subsection{End-to-end Learning for Uncorrelated Feature Encoding}

Our style transfer network has an architecture of \{encoder, feature transformer, decoder\} as shown in fig.\ref{fig:network_architecture}. Encoder and decoder networks have the same structure of VGG\cite{Simonyan14c} feature network and the mirrored VGG structure respectively as the previous works\cite{Huang_2017_ICCV,Li_2017_NIPS}. As the feature transformer, AdaIN layer\cite{Huang_2017_ICCV} is used to adjust the encoded feature $x_{c}$ distribution of content image to the encoded feature $x_{s}$ distribution of style image without considering inter-channel correlation.

For the objective function of network training, we use the weighted combination of content, style, and uncorrelation losses as Eq.\ref{eq:total_loss}. Here, L2 loss between the encoded features of output and content images is used as the content loss like the previous feed-forward network learning\cite{Gatys_2016_CVPR,Johnson_2016_ECCV,Huang_2017_ICCV} and L2 loss between Gram matrices of output and target style images is used as the style loss like the original neural style method\cite{Gatys_2016_CVPR} to consider inter-channel correlation in feature distribution. Uncorrelation loss is calculated as Eq.\ref{eq:uncorrelation_loss}. Using appropriate weights for style loss and uncorrelation loss is essential for successful training because, with large values of weights, the content of encoded feature can be eliminated while the style expression on the output image will be strong and the inter-channel correlation of encoded feature will be removed completely. Therefore, the weight values of the losses ($\lambda_{c}$, $\lambda_{s}$, $\lambda_{r}$) are experimentally determined as ($1.0$, $50.0$, $0.01$).

\begin{equation} \label{eq:total_loss}
\mathcal{L}_{total} = \lambda_{c}\mathcal{L}_{content} + \lambda_{s}\mathcal{L}_{style} + \lambda_{r}\mathcal{L}_{uncorrelation}
\end{equation}

\subsection{Cascade Network Scheme for Multiple Scaled Style Transfer}

\begin{figure}[t]
\begin{center}
\includegraphics[trim={0cm 4.3cm 0cm 6cm},clip,width=1.0\linewidth]{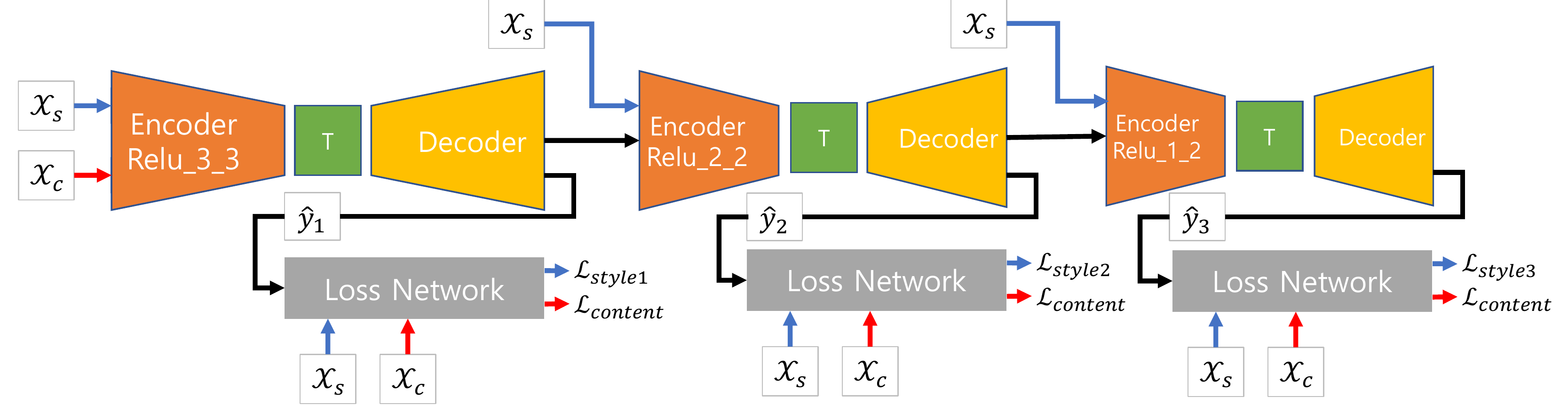}
\end{center}
   \caption{Our uncorrelated feature encoding method can be also used with the cascade network scheme of \cite{Li_2017_NIPS} to transform multiple scales of style independently. The first network takes original content image, and the second and third networks use the image generated from the previous network as its content input.}
\label{fig:network_cascade}
\end{figure}

\begin{figure}[t]
\begin{center}
\includegraphics[trim={0cm 1cm 0cm 0cm},clip,width=1.0\linewidth]{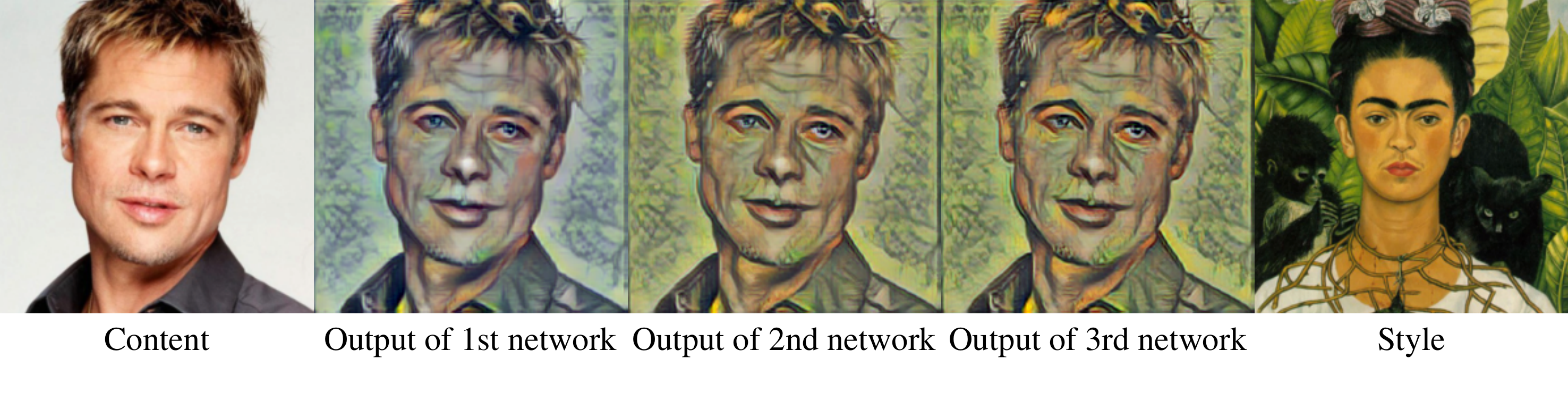}
\end{center}
   \caption{These images show the output stylized images of the cascade networks. As the content image goes through the cascade networks, small patterns appear on the stylized image and color tone of the stylized image becomes similar to the style image.}
\label{fig:cascade_images}
\end{figure}

The spatial and abstract information of the encoded feature differs according to the hierarchical level of layers in a convolutional neural network \cite{Gatys_2016_CVPR,Johnson_2016_ECCV}. For this reason, Li et al.\cite{Li_2017_NIPS} proposed a multiple scaled style transferring scheme using cascade networks. Our uncorrelation feature encoding method can also cooperate with this cascade scheme to improve the quality of style transfer. Our cascade networks consist of three networks as shown in fig.\ref{fig:network_cascade}. Each network has encoder and decoder networks according to one scale of style and trained with the same content loss at the content layer and the accumulated style loss up to the corresponding scale.

Fig.\ref{fig:cascade_images} shows an example of output stylized images through the cascade of three networks. As the image goes through the cascade, small patterns appear and color tone is corrected on the output image.

\section{Experimental Results}

\subsection{Experimental Setup}

As the first feed-forward network approach\cite{Johnson_2016_ECCV}, we used response of \{$relu3\_3$\} as the content feature and the responses of \{$relu1\_2$, $relu2\_2$, $relu3\_3$, $relu4\_3$\} as the style features. According to this content and style configuration, the structure of VGG16 feature network up to \{$relu3\_3$\} layer was used for encoder network and its mirrored structure was use for decoder network.

For the training dataset, we used MS-COCO\cite{Lin_2014_ECCV} for content images and Painter By Number\cite{painter_by_numbers} for style images. Each dataset consists of approximately 80,000 images. As a pre-processing, the images were resized to 256 pixels maintaining the aspect ratio and then randomly cropped to 240 pixels to avoid boundary artifact and to augmenting training data. With those pre-processed images, our network was trained by using Adam optimizer\cite{Kingma_2015_ICLR} with training parameters of 4 epochs, 8 random \{content image, style image\} pairs per batch, and learning rates $10^{-4}$ for decoder network. For encoder network, we used smaller learning rate $10^{-5}$ for fine-tunning of the encoder parameters from the initial pre-trained VGG16 parameter values.

\subsection{Uncorrelated Feature Encoding}

\begin{figure}[t]
\begin{center}
\includegraphics[trim={0cm 0cm 0cm 0cm},clip,width=1.0\linewidth]{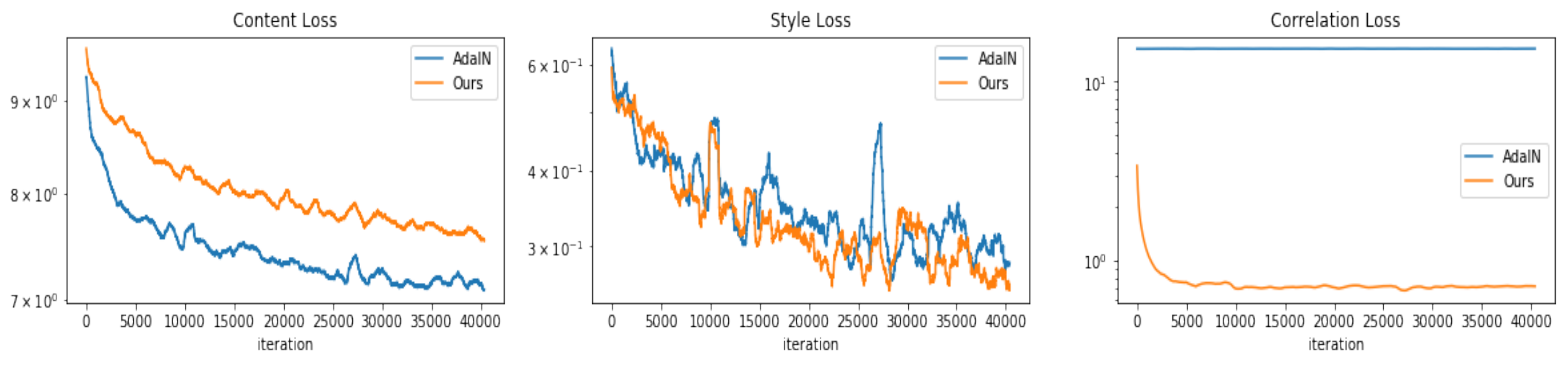}
\end{center}
   \caption{Loss reduction in training process: Each loss is stored for every iteration with batch size 8 during 4 epochs and the average values with 100-iteration mask are used to plot these graphs to reduce fluctuating noise in loss. It is possible to encode an uncorrelated feature with our encoder network trained to minimize uncorrelation loss (right) without much degradation in content (left) and style (center) losses compared to AdaIN \cite{Huang_2017_ICCV} with Gram style loss and fixed encoder network.}
\label{fig:losses}
\end{figure}

\begin{figure}[t]
\begin{center}
\includegraphics[trim={0cm 12cm 0cm 0cm},clip,width=0.9\linewidth]{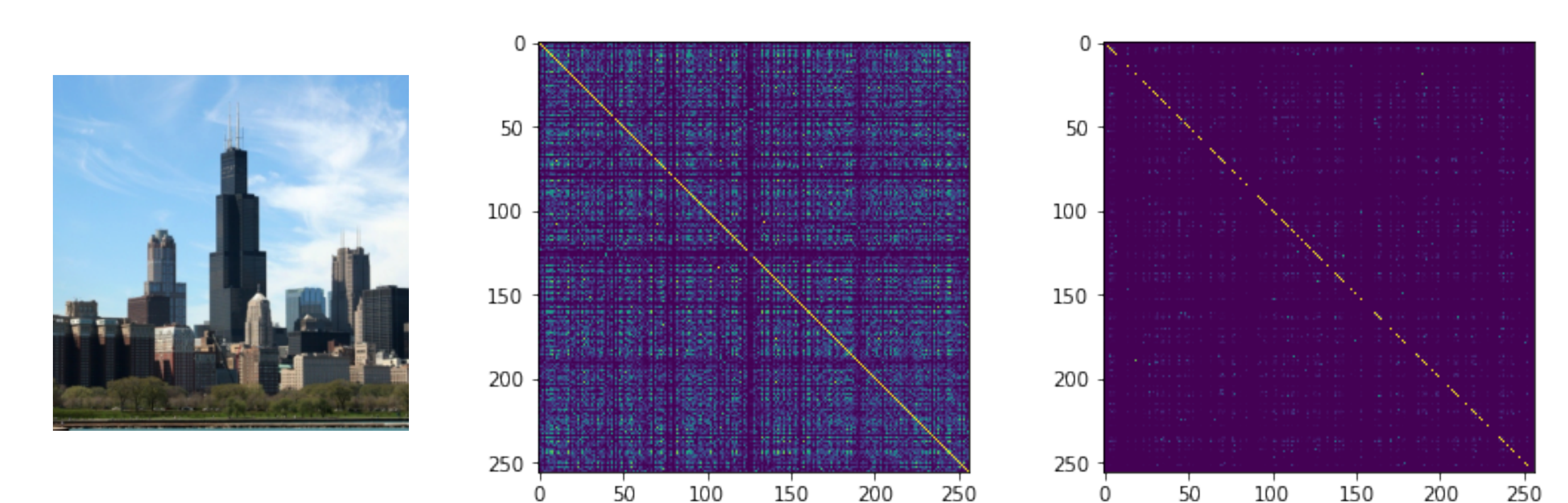}
\end{center}
   \caption{A sample content image (left), correlation coefficient matrices of the extracted features from initial VGG16 feature network (center) and from our trained encoder network (right): After training encoder network with uncorrelation loss (eq.\ref{eq:uncorrelation_loss}, the inter-channel correlation (off-diagonal pixels of the center image) in the encoded feature is almost diminished (off-diagonal pixels of the right image).}
\label{fig:cosine matrix}
\end{figure}

Fig.\ref{fig:losses} shows the reduction of content, style, and uncorrelation losses during training process compared to the training result of AdaIN\cite{Huang_2017_ICCV} with Gram style loss and fixed encoder network. As the training iteration goes, our uncorrelation loss fastly reduced to $0.7$ which is only about $4\%$ of the initial uncorrelation loss $17.0$. Fig.\ref{fig:cosine matrix} shows an example content image (left) and two correlation coefficient matrices of its encoded features extracted by the initial VGG encoder network (center) and our trained encoder network (right). Each pixel value of the matrices represents correlation coefficient (eq.\ref{eq:cosine_distance}) between two channels corresponding to the coordinate. We can find that the encoded feature with our trained encoder has very low inter-channel correlation (dark off-diagonal pixels) while that of initial VGG encoder has high inter-channel correlation (bright off-diagonal pixels).

When inter-channel correlation was removed from the encoded feature, there were no change in the final style loss (center image of fig.\ref{fig:losses}) and $5\%$ increase in the final content loss (left image of fig.\ref{fig:losses}) after 40,000 iterations. This slight degradation in content loss did not affect to the quality of output stylized images. We will compare the image qualities of our method and the previous methods in the later subsection.

\begin{figure}[t]
\begin{center}

\includegraphics[trim={0cm 0cm 0cm 0cm},clip,width=1.0\linewidth]{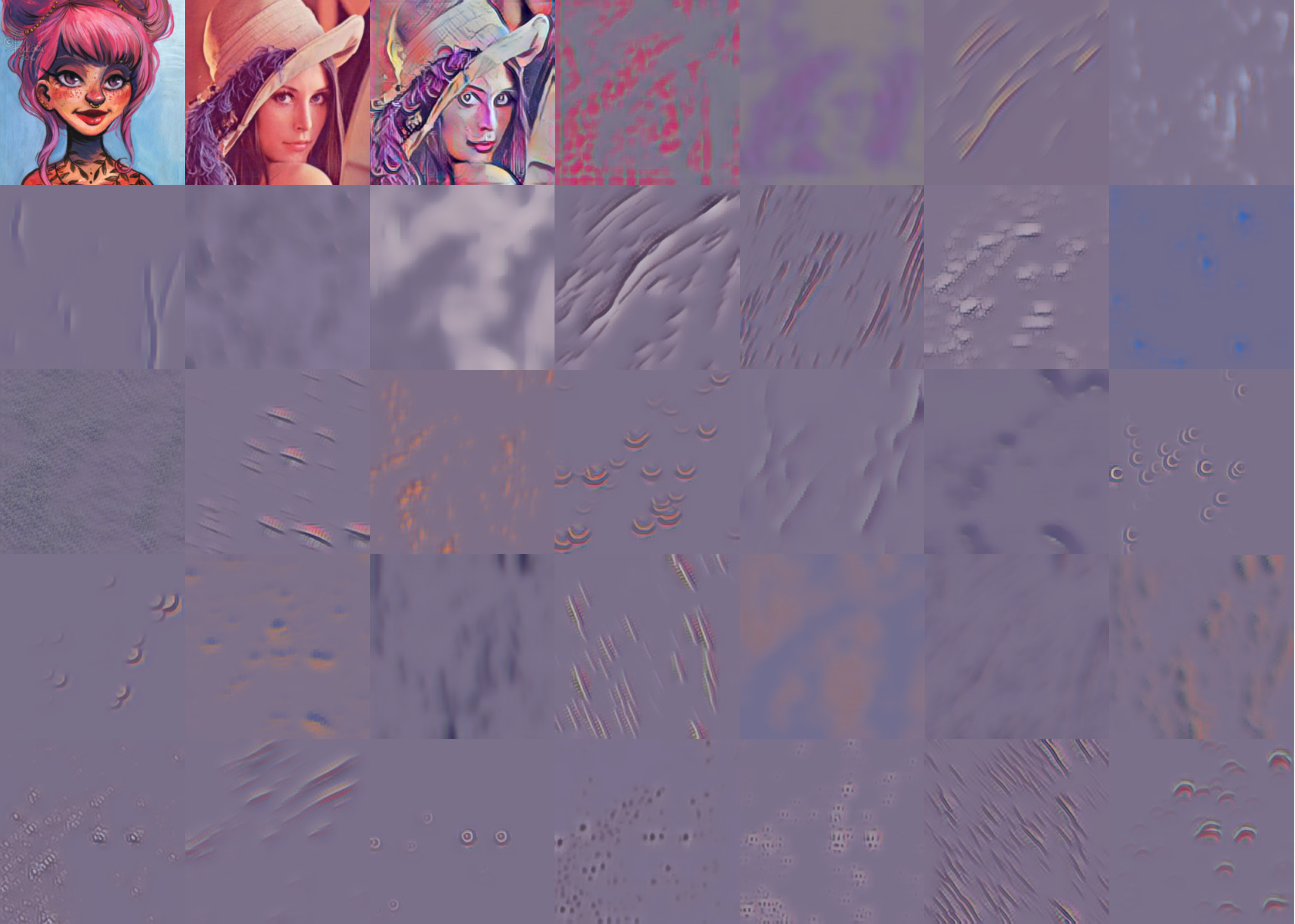}
.
\end{center}
   \caption{From left to right on the first row, style image, content image, style transferred image, and the decoded images of each independent channel of the stylized feature map. As shown in the decoded images, each channel represents independent image features like edge, texture, and color.}
\label{fig:channel_response}
\end{figure}

The decoded channels of the stylized feature of another sample content and target style pair are presented in fig.\ref{fig:channel_response}. We can find that each channel represents different shape or texture or color as an independent component of the output stylized image.

\subsection{Redundant Channel Elimination}

During the network training process with our uncorrelation loss, we observed that not only inter-channel correlation but also intra-channel variance was reduced. The sparseness of the diagonal pixels of a sample correlation coefficient matrix on the right image of fig.\ref{fig:cosine matrix} shows an example of this situation. This can be understood that redundant channels were eliminated as the channels were getting orthogonalized to be independent of each other.

\begin{figure}[t]
\begin{center}
\includegraphics[trim={0cm 0cm 0cm 0cm},clip,width=1.0\linewidth]{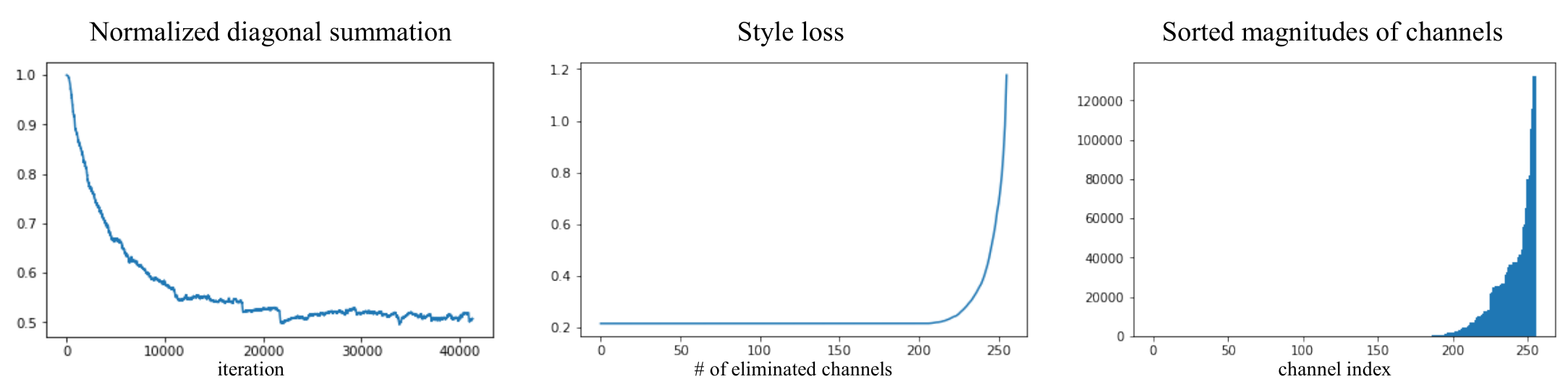}
\end{center}
   \caption{Change of normalized diagonal summation during training process (left), Change of style loss as the number of eliminated channels increases (center), Sorted channel magnitudes of a sample encoded feature of the style image on fig.\ref{fig:channel_eliminated_images} (right)}
\label{fig:diagonal_analysis}
\end{figure}

To measure how many channels are eliminated during the training process, we drew the normalized diagonal summation (eq.\ref{eq:diagonal_value}) of the correlation coefficient matrices of batch data. Because each diagonal pixel value of the correlation coefficient matrix is one for non-zero channel or zero for zero channel, the normalized diagonal summation represents the ratio of the number of non-zero channels to total number of channels.
\begin{equation} \label{eq:diagonal_value}
D = \frac{1}{2C} (\sum_{i=1}^{C}{r_c({f_c}_i, {f_c}_i)} + \sum_{i=1}^{C}{r_c({f_s}_i, {f_s}_i)})
\end{equation}
As shown in the left image of fig.\ref{fig:diagonal_analysis}, the normalized diagonal summation is reduced to $0.5$. This means the number of non-zero channels was reduced to $50\%$ of the total channels after training process.

\begin{figure}[t]
\begin{center}
\includegraphics[trim={0cm 1cm 0cm 1cm},clip,width=1.0\linewidth]{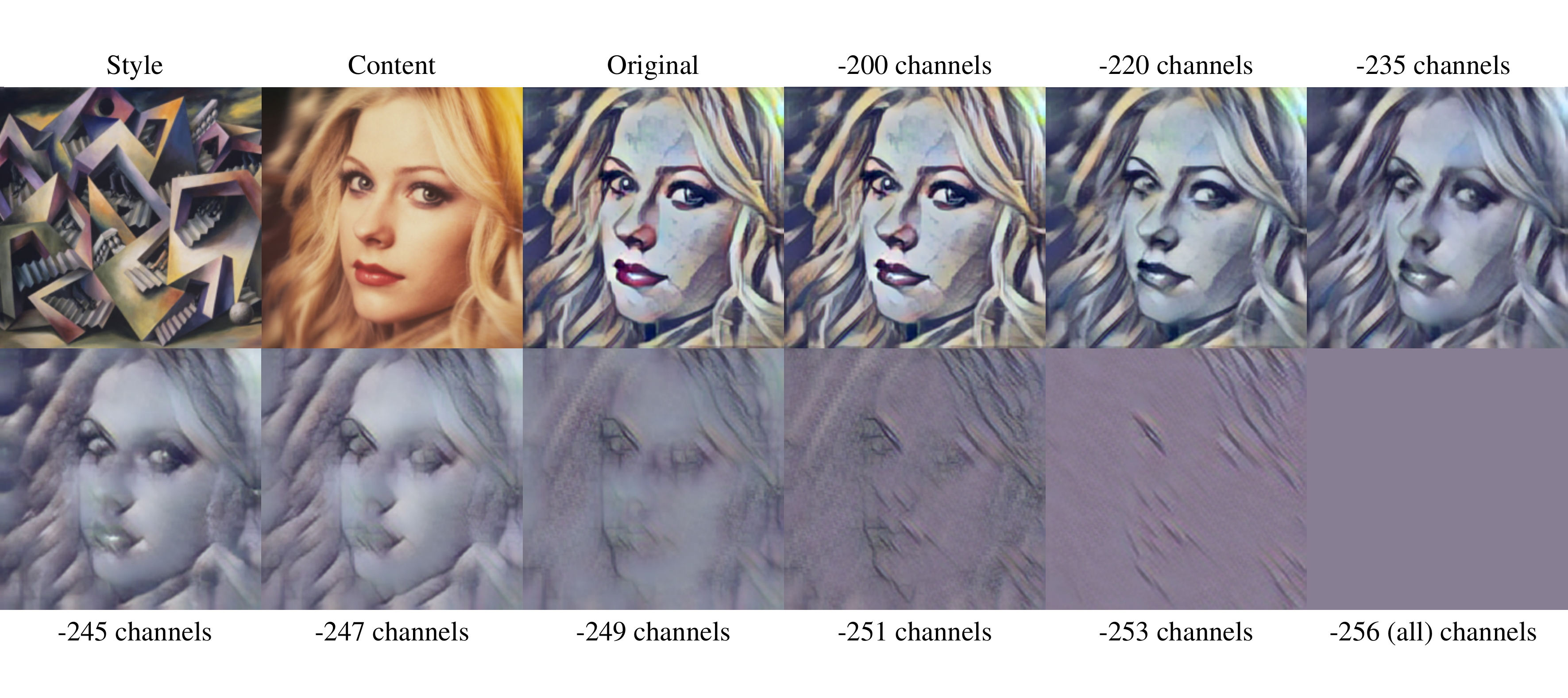}
\end{center}
   \caption{Decoded images of encoded feature with channel elimination: The style quality does not change until eliminating 200 channels of smallest magnitudes (-200 channels). As the number of eliminated channels increases from 200 to 256 (all), components of image like color and shape shown in fig.\ref{fig:channel_response} disappear from the decoded image.}
\label{fig:channel_eliminated_images}
\end{figure}

The right image of fig.\ref{fig:diagonal_analysis} shows the ordered magnitudes of the encoded feature channels of the style image on fig.\ref{fig:channel_eliminated_images}. Here, we can also find that a large portion of channels are redundant. The center image of fig.\ref{fig:diagonal_analysis} shows how the style loss is changing as the number of removed channels is increasing. The channels of smaller magnitude were eliminated first. Up to 200 eliminated channels, there is almost no change in style loss in the center image of fig.\ref{fig:diagonal_analysis} and no degradation in the output stylized image as shown in fig.\ref{fig:channel_eliminated_images}. For numbers of eliminated channels larger than 200, the style loss increase quickly (fig.\ref{fig:diagonal_analysis} and the output stylized image became lack of content and style (fig.\ref{fig:channel_eliminated_images}). Because the magnitudes of encoded feature channels change image to image, the indices of redundant channels differ depending on image. Therefore, we experimentally decided to keep $80\%$ of total magnitude of channels and this setting was used for further experiments.

\subsection{Comparing to the Previous Methods}

\subsubsection{Quality of Stylized Images}

\begin{figure}[t]
\begin{center}
\includegraphics[trim={0cm 0cm 0cm 0cm},clip,width=1.0\linewidth]{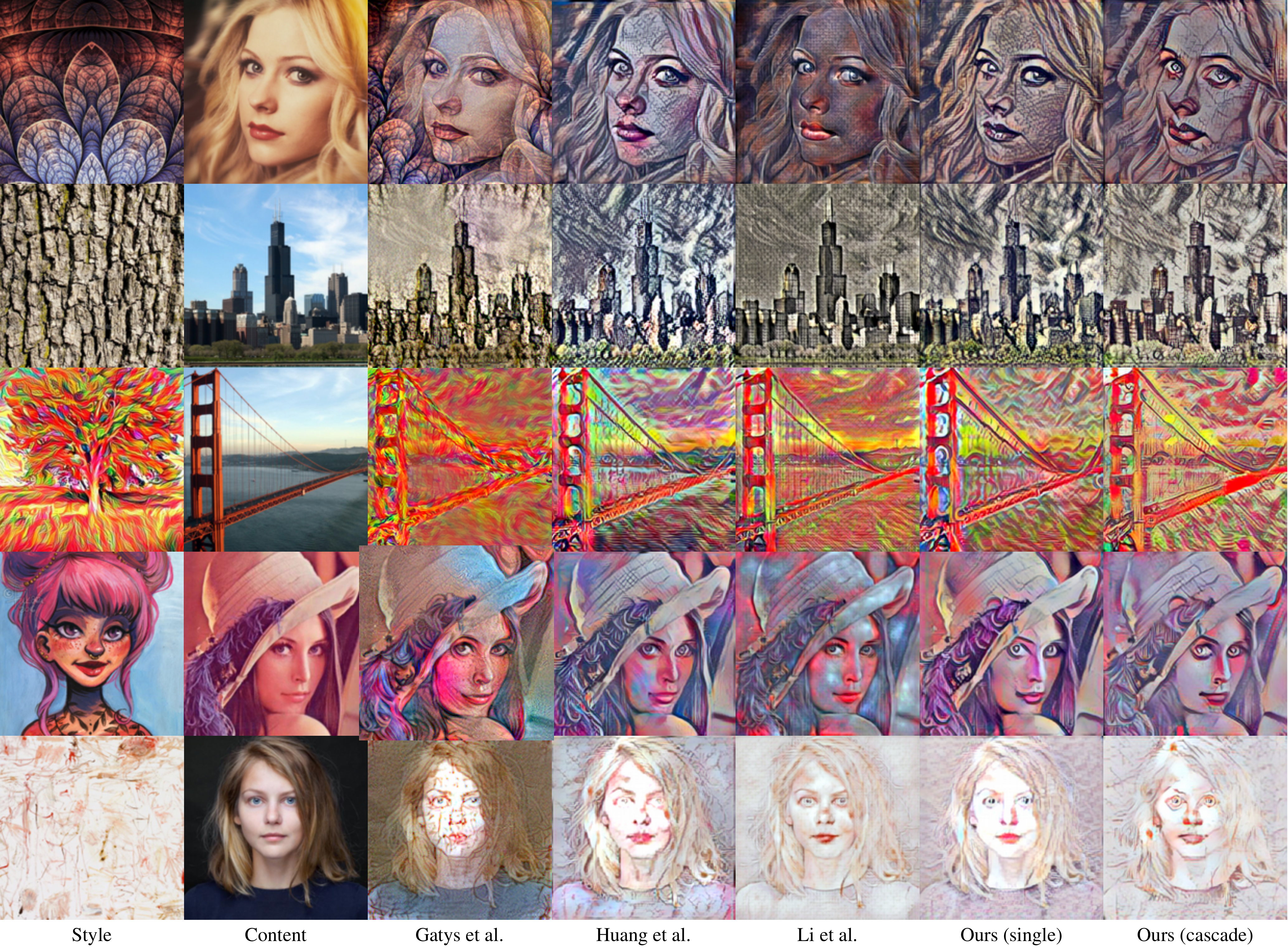}
\end{center}
   \caption{Examples of output stylized images with serveral image style transfer methods: The target style images on the first column are not used in network training.}
\label{fig:methods_comparison}
\end{figure}

We compared the output stylized images of our method with single network or with cascade of three networks, neural style\cite{Gatys_2016_CVPR} with 500 iterations, AdaIN\cite{Huang_2017_ICCV} with VGG16 and the same network structure of ours, and WCT\cite{Li_2017_NIPS} with VGG16 and three cascade networks. All networks were trained with the same setup described in the section of Experimental Setup.

The output stylized images for several pairs of content and target style images are drawn in fig.\ref{fig:methods_comparison}. As the other two feed-forward network methods\cite{Huang_2017_ICCV,Li_2017_NIPS}, our methods with single network or cascade networks also have some degradation in style quality of the output images compared to neural style\cite{Gatys_2016_CVPR}. However, thanks to the fine-tunned encoder network for style transfer, the result images of our method with single network are slightly better than those of \cite{Huang_2017_ICCV} in color tone and detailed texture by using the same network structure. Compared to the results of \cite{Li_2017_NIPS}, our method with cascade scheme generated the stylized images of various patterns and tones more similar to the style image. Because our method used style loss to train decoder network while Li et al.\cite{Li_2017_NIPS} used reconstruction loss, it seems that our decoder network generated the stylized images more similar to the target style.

\subsubsection{Image Generating Speed}

\begin{table}[t] \centering 
\caption{Comparison of style transferring methods: The elapsed times are average values of 1,000 trials and written in milliseconds.}
\begin{tabular}{l c c c c c}

Method     & \thead{time \\ 240 pixels}   & \thead{time \\ 512 pixels} & \thead{correlation \\ in feature} & \thead{correlation \\ in transformer} & \thead{multi-scaled \\ style} \\
\hline
Gatys et al.\cite{Gatys_2016_CVPR} & 5210 & 18400 & - & - &\\

Huang et al.\cite{Huang_2017_ICCV} & 8.75 & 34.70 & Yes & No & No\\

Li et al.\cite{Li_2017_NIPS} & 62.10 & 104.00 & Yes & Yes & Yes\\

Ours (single) & 7.41 & 28.80 & No & No & No\\

Ours (cascade) & 12.00 & 45.20 & No & No & Yes\\

\label{table:speed}	
\end{tabular}
\end{table}

We measured the elapsed times of forward style transfer processing with our method and the previous methods by calculating the average duration of 1,000 trials. Those methods were performed on Pytorch framework with CUDA 9.0, CuDNN 7.0, and NVIDIA 1080 Ti GPU card. Table.\ref{table:speed} shows the measured elapsed times. Both with 240 pixel sized image and with 512 pixel sized image, our method achieved the fastest forward processing speed. Because of using the light-weighted uncorrelated feature transform (AdaIN) of \cite{Huang_2017_ICCV}, our method with cascade networks shows much faster forward processing speed than Li el al.\cite{Li_2017_NIPS} with heavy correlated feature transform (WCT). Compared to Huang et al.\cite{Huang_2017_ICCV}, our method is slightly faster because channel elimination was done on both side of adjacent layers of encode and decoder networks in the same network structure of \cite{Huang_2017_ICCV}.

\subsection{Style Strength Control}

\begin{figure}[t]
\begin{center}
\includegraphics[trim={2cm 6.5cm 2cm 7cm},clip,width=1.0\linewidth]{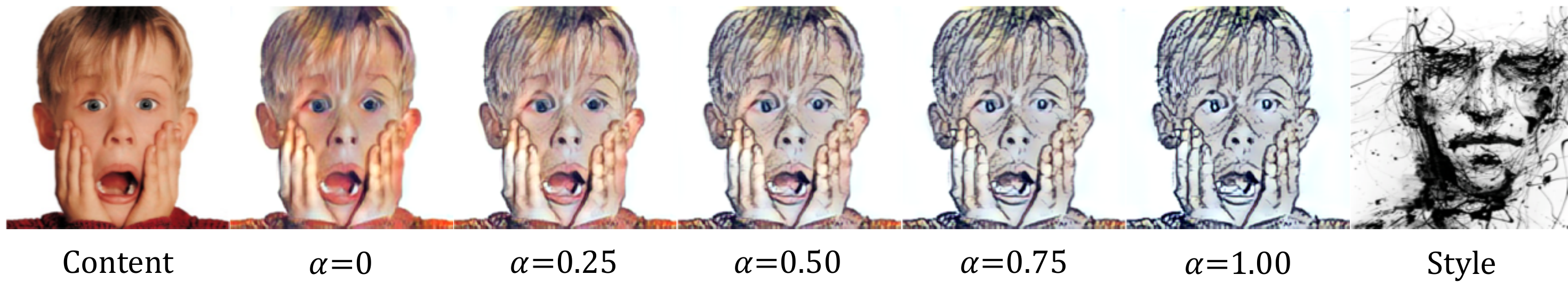}
\end{center}
   \caption{Run time user control of style strength}
\label{fig:alpha_control}
\end{figure}

Our method can control the style strength of the output image with user control parameter $\alpha$ at run time as the same manner of the previous fast style transfer methods\cite{Huang_2017_ICCV,Li_2017_NIPS,Li_2017_IJCAI}. Fig.\ref{fig:alpha_control} shows an example result of style strength control.

\section{Conclusion}

In this work, we proposed an end-to-end network learning method for image style transferring with uncorrelated feature encoding. Based on the experimental result, it was shown that our uncorrelation loss is effective in training the encoder network to generate uncorrelated feature without loss of content and style representing ability. The independent channels of the encoded feature allowed the feature transformer of style transfer network to use AdaIN method of much lower computational cost for correlation-existing style than WCT with heavy calculation of inverse square root of covariance matrices, and this resulted in faster forward processing speed than the previous correlation-considering method. As an another effect of encoder network learning, it was experimentally proven that a large portion of the encoded feature channels is redundant and, therefore, can be removed without degradation in style loss and image quality.

Although our uncorrelation feature encoding technique was applied only to the content feature layer in this work, it can be applied to multiple scales of feature layers without modification. So, we are expecting much more channel reduction in multiple scales of feature layers and this will achieve much faster forward processing speed. And we can consider another loss which forces the encoder network to reduce the number of encoded feature channels directly in consistent indices. Uncorrelated feature encoding in multiple scales of feature layers and forcing further reduction of channels will prune the network efficiently for much faster real-time process. This is remained for our future works.

\clearpage

\bibliographystyle{splncs}
\bibliography{egbib}

\begin{thebibliography}{10}

\bibitem{Huang_2017_ICCV}
Huang, X., Belongie, S.:
\newblock Arbitrary style transfer in real-time with adaptive instance
  normalization.
\newblock In: The IEEE International Conference on Computer Vision (ICCV). (Oct
  2017)

\bibitem{Simonyan14c}
Simonyan, K., Zisserman, A.:
\newblock Very deep convolutional networks for large-scale image recognition.
\newblock CoRR \textbf{abs/1409.1556} (2014)

\bibitem{Johnson_2016_ECCV}
Johnson, J., Alahi, A., Fei-Fei, L.:
\newblock Perceptual losses for real-time style transfer and super-resolution.
\newblock In: European Conference on Computer Vision (ECCV). (Oct 2016)

\bibitem{Gatys_2016_CVPR}
Gatys, L.A., Ecker, A.S., Bethge, M.:
\newblock Image style transfer using convolutional neural networks.
\newblock In: The IEEE Conference on Computer Vision and Pattern Recognition
  (CVPR). (June 2016)

\bibitem{Ulyanov_2016_ICML}
Ulyanov, D., Lebedev, V., Vedaldi, A., Lempitsky, V.S.:
\newblock Texture networks: Feed-forward synthesis of textures and stylized
  images.
\newblock In: International Conference on Machine Learning (ICML). (June 2016)

\bibitem{Dumoulin_2017_ICLR}
Dumoulin, V., Shlens, J., Kudlur, M.:
\newblock A learned representation for artistic style.
\newblock In: International Conference on Learning Representations (ICLR). (Apr
  2017)

\bibitem{Li_2017_NIPS}
Li, Y., Fang, C., Yang, J., Wang, Z., Lu, X., Yang, M.H.:
\newblock Universal style transfer via feature transforms.
\newblock In: Advances in Neural Information Processing Systems. (2017)
  385--395

\bibitem{Ghiasi_2017_BMVC}
Ghiasi, G., Lee, H., Kudlur, M., Dumoulin, V., Shlens, J.:
\newblock Exploring the structure of a real-time, arbitrary neural artistic
  stylization network.
\newblock In: British Machine Vision Conference (BMVC). (Sep 2017)

\bibitem{Ulyanov_2017_CVPR}
Ulyanov, D., Vedaldi, A., Lempitsky, V.:
\newblock Improved texture networks: Maximizing quality and diversity in
  feed-forward stylization and texture synthesis.
\newblock In: The IEEE Conference on Computer Vision and Pattern Recognition
  (CVPR). (July 2017)

\bibitem{Gatys_2017_CVPR}
Gatys, L.A., Ecker, A.S., Bethge, M., Hertzmann, A., Shechtman, E.:
\newblock Controlling perceptual factors in neural style transfer.
\newblock In: The IEEE Conference on Computer Vision and Pattern Recognition
  (CVPR). (July 2017)

\bibitem{Luan_2017_CVPR}
Luan, F., Paris, S., Shechtman, E., Bala, K.:
\newblock Deep photo style transfer.
\newblock In: The IEEE Conference on Computer Vision and Pattern Recognition
  (CVPR). (July 2017)

\bibitem{Li_2017_IJCAI}
Li, Y., Wang, N., Liu, J., Hou, X.:
\newblock Demystifying neural style transfer.
\newblock In: IJCAI. (2017)

\bibitem{Sun_2016_AAAI}
Sun, B., Feng, J., Saenko, K.:
\newblock Return of frustratingly easy domain adaptation.
\newblock In: Proceedings of the Thirtieth AAAI Conference on Artificial
  Intelligence. AAAI'16, AAAI Press (2016)  2058--2065

\bibitem{Sun_2016_ECCV}
Sun, B., Saenko, K.:
\newblock Deep coral: Correlation alignment for deep domain adaptation.
\newblock In: ECCV 2016 Workshops. (2016)

\bibitem{Sedgwicke4483}
Sedgwick, P.:
\newblock Pearson's correlation coefficient.
\newblock BMJ \textbf{345} (2012)

\bibitem{Lin_2014_ECCV}
Lin, T.Y., Maire, M., Belongie, S., Hays, J., Perona, P., Ramanan, D., Dollár,
  P., Zitnick, C.L.:
\newblock Microsoft {COCO:} common objects in context.
\newblock In: European Conference on Computer Vision (ECCV). (Sep 2014)

\bibitem{painter_by_numbers}
Nichol, K.:
\newblock Kaggle dataset: Painter by numbers (2016)
  \url{https://www.kaggle.com/c/painter-by-numbers}.

\bibitem{Kingma_2015_ICLR}
Kingma, D.P., Ba, J.:
\newblock Adam: {A} method for stochastic optimization.
\newblock In: International Conference on Learning Representations (ICLR). (May
  2015)

\end{thebibliography}
\end{document}